\documentclass[review]{elsarticle}

\journal{}

\usepackage{subfig}

\usepackage{amssymb}
\usepackage{amsmath}
\usepackage{siunitx}

\usepackage[linesnumbered,ruled]{algorithm2e}

\usepackage{numcompress}\biboptions{authoryear}

\makeatletter
\def\ps@pprintTitle{%
 \let\@oddhead\@empty
 \let\@evenhead\@empty
 \def\@oddfoot{\centerline{\thepage}}%
 \let\@evenfoot\@oddfoot}
\makeatother

\begin{document}

\begin{frontmatter}

\title{Fast and Accurate Tumor Segmentation of Histology Images using Persistent Homology and Deep Convolutional Features
}



\author[Warwick]{Talha Qaiser}
\address[Warwick]{Department of Computer Science, University of Warwick, UK}
\address[UHCW]{Department of Pathology, University Hospitals Coventry and Warwickshire, UK}

\author[UHCW]{Yee-Wah Tsang}

\author[hiroshima]{Daiki Taniyama}

\author[hiroshima]{Naoya Sakamoto}
\address[hiroshima]{Department of Molecular Pathology, Hiroshima University Institute of Biomedical and Health Sciences, Japan}

\author[osaka]{Kazuaki Nakane}

\address[osaka]{Graduate School of Medicine, Division of Health Science, 
Osaka University, Japan}

\author[maths]{David Epstein}
\address[maths]{Mathematics Institute, University of Warwick, UK.}

\author[Warwick,UHCW,ATI]{Nasir Rajpoot}
\address[ATI] {The Alan Turing Institute, UK}

\begin{abstract}Tumor segmentation in whole-slide images of histology slides is an important step towards computer-assisted diagnosis. In this work, we propose a tumor segmentation framework based on the novel concept of \textit{persistent homology profiles} (PHPs). For a given image patch, the homology profiles are derived by efficient computation of persistent homology, which is an algebraic tool from homology theory. We propose an efficient way of computing topological persistence of an image, alternative to simplicial homology. The PHPs are devised to distinguish tumor regions from their normal counterparts by modeling the atypical characteristics of tumor nuclei. We propose two variants of our method for tumor segmentation: one that targets speed without compromising accuracy and the other that targets higher accuracy. The fast version is based on selection of exemplar image patches from a convolution neural network (CNN) and  patch classification by quantifying the divergence between the PHPs of exemplars and the input image patch. Detailed comparative evaluation shows that the proposed algorithm is significantly faster than competing algorithms while achieving comparable results. The accurate version combines the PHPs and high-level CNN features and employs a multi-stage ensemble strategy for image patch labeling. Experimental results demonstrate that the combination of PHPs and CNN features outperforms competing algorithms. This study is performed on two independently collected colorectal datasets containing adenoma, adenocarcinoma, signet and healthy cases. Collectively, the accurate tumor segmentation produces the highest average patch-level F1-score, as compared with competing algorithms, on malignant and healthy cases from both the datasets. Overall the proposed framework highlights the utility of persistent homology for histopathology image analysis. 

\end{abstract}


\end{frontmatter}


\section{Introduction}
\label{introduction}

Colorectal cancer (CRC), also known as colon or  bowel cancer, originates in the colon or the  rectum, due to abnormal growth pattern of cells. CRC is the third most commonly diagnosed cancer in males and the second most in females \cite{ferlay2010estimates}, with an estimated 1.4 millions cases and 693,000 deaths occurring in 2012 \cite{torre2015global}. According to the National Cancer Institute (NCI), around $4.3\%$ of the human population will be diagnosed with CRC during their lifetime, based on 2012-2014 data\footnote{https://seer.cancer.gov/statfacts/html/colorect.html}. CRC results from excessive growth of malignant (cancer) cells in the colon or the rectum. The most common form of CRC is adenocarcinoma (found in upto 95$\%$ of CRC cases) which develops in epithelial gland cells--- these are the glands that secrete mucus, which lubricates the colorectral region. 

In a routine diagnostic process, a pathologist analyzes tissue sections on glass slides under the microscope to observe morphological features and variability of nuclear morphology. However, careful visual examination of tissue slides is difficult when workloads are high, and the subjective nature of the task inevitably leads
to inter and even intra-observer variability \cite{viray2013prospective, smits2014estimation}. In contrast, automated algorithms can analyze the data with reproducible results. The reliability of the results of an algorithm can be objectively measured (for example against a patient's subsequent clinical progress) and then improved against an objective standard. This is now possible with the advent of digital slides scanners, as hundreds of glass tissue slides can now be scanned in a single run of the scanner. A whole-slide image (WSI) is a multi-resolution gigapixel image typically stored in a pyramid structure, formed by scanning a conventional glass slide at microscopic resolution. In view of the increasing number of CRC cases and shortcomings of the conventional diagnosis system, it is imperative to develop fast and reliable algorithms that can assist the histopathologists in their diagnosis of cancer.     

Localization of malignant tumor regions in Hematoxylin and Eosin (H\&E) stained slides is an important first task for a pathologist while diagnosing CRC. Manual segmentation of tumor regions from glass slides is a challenging and time consuming task. Therefore, automated localization  of tumor-rich areas is a vital step towards a computer-assisted diagnosis and quantitative image analysis. Accurate segmentation of tumor-rich areas may also assist pathologists in understanding disease aggressiveness and selection of high power fields for tumor proliferation grading and scoring. In a recent study \cite{dunne2016challenging}, it has been shown that precise localization of tumor epithelial regions in CRC can overcome the association of nonmalignant stroma regions in gene expression profiling, which also provides substantial prognostic information for individual cases. Hence, automated tumor segmentation of CRC tissue slides could potentially speed up the diagnostic process and overcome the inter-observer variability of conventional methods \cite{litjens2016deep, HIS:HIS13333}. 

\begin{figure}[t]
\centering
\includegraphics[width=1.0\textwidth]{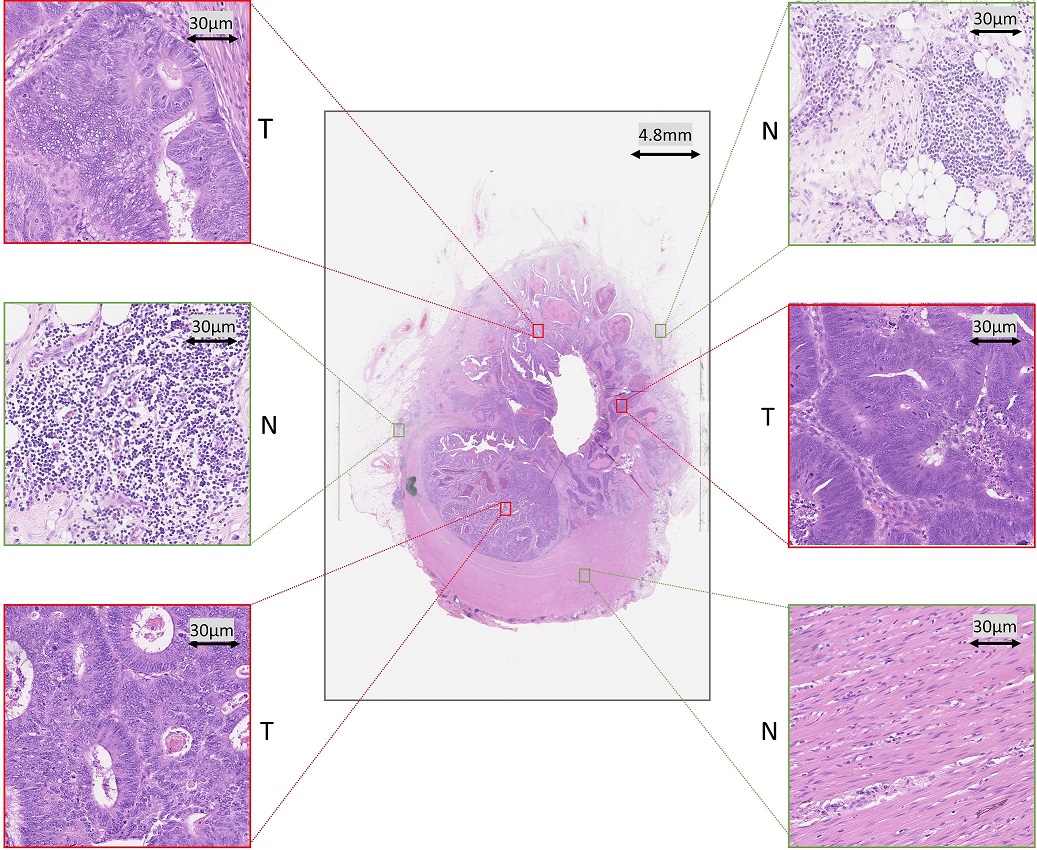}
\caption{An example of a whole slide image with 6 regions-of-interest (ROIs) to illustrate the degree of connectivity between nuclei in tumor and normal regions. The zoomed-in regions are of size 140.8 $\times$ 140.8  $\mu m^2$, which is equivalent to 20$\times$ magnification. ROIs with green rectangles (with label N) shows non-tumor whereas subregion with red rectangles (with label T) shows tumor areas.}
\label{fig:cell_structure}
\end{figure}

Tumor regions can be distinguished from normal regions using the appearance of cell nuclei \cite{baba2007comparative, clevert2015fast}. In tumor regions epithelial nuclei have atypical characteristics ---   relatively large nuclei, with heterogeneous chromatin texture and irregularities in their shape and size. Due to uncontrolled cell division, tumor nuclei sometimes form clusters filling inter-cellular regions, as shown in Figure \ref{fig:cell_structure}. In some cases with moderately and poorly differentiated grades, the structure of individual nuclei is difficult to discern. In contrast, nuclei retain their structure and morphological appearance in normal regions including stroma, lymphocytes, normal mucosa and adipose tissue regions.

In this paper, we show that important morphological differences between normal and cancer nuclei can be measured using persistent homology, a mathematical tool explained in Section~\ref{PHP}. We propose two persistent homology methods for tumor segmentation of H\&E stained WSIs including \textit{a)} homology based fast and reliable tumor segmentation \textit{b)} accurate tumor segmentation by combining the homological and deep convolutional features to enhance the classification accuracy of a deep convolutional neural networks (CNN).

We validate the proposed methods with relatively large datasets containing both malignant and healthy cases from two independent institutions. We repeat results of weak and strong cross-validation on the two datasets. Generally in a clinical setup, a WSI scanner processes 500 to 1000 glass slides each day and so analyzing data from a single scanner may require the processing of several terabytes of new data each day. We therefore make a particular point in this paper of quoting run times, and show that our fast and reliable tumor segmentation algorithm is significantly faster than a conventional convolution neural network (CNN) and other competing approaches.

\subsection{Our Contributions}
We start with an algorithm to derive \textit{persistent homology profiles} which effectively transform an input patch into two 1-D statistical distributions. These  distributions capture the degree of nuclear connectivity in a given patch of a WSI. We propose two approaches for tumor segmentation based on the PHPs. 

In the first instance, we seek to develop a fast and reliable tumor segmentation algorithm. Presented with an image patch from a WSI, the algorithm computes the PHP, and
then compares it with PHPs derived from tumor and normal \textit{exemplar patches} using the symmetrized Kullback-Leibler divergence (KLD). The exemplar patches are precomputed, using CNN activations but the CNN itself is not used again after precomputation. Finally, we employ a novel variant of $k$-NN classifier to predict the outcome of a given patch as tumor or non-tumor. The fast algorithm, compared with the state-of-the-art, is significantly faster while competitive in accuracy.

Our second algorithm for accurate tumor segmentation is a combination of deep convolutional and topological features (PHPs). In contrast to the fast algorithm, this algorithm continues to use a CNN and extracts a set of learned features. We compute PHPs of the training dataset and separately employ a Random Forest regression model for both set of features. We propose a multi-stage ensemble strategy to fuse the output from regression models. Experimental results on both the datasets demonstrate that the combination of topological and CNN features produces high segmentation accuracy.  

Preliminary versions of these approaches were presented in \cite{qaiser2016persistent, qaiser2017tumor}. In this paper, we provide a comprehensive literature review on tumor segmentation of histology WSIs, an intuitive explanation of persistent homology and our proposed algorithms by providing more details on the underlying principles. We report the results of our methods on two independent datasets comprising a range of malignant CRC cases including adenoma, adenocarcinoma, signet as well as healthy cases. This paper also contains a more robust approach for selection of exemplar patches from a CNN, discussion and comparative results on alternative options for selection of exemplar patches, runtime analysis of tumor segmentation approaches and finally, a detailed analysis on robustness of the proposed methods. 

\subsection{Overview}
The rest of the article is organized as follows: In Section ~\ref{related work}, we briefly review some notable tumor segmentation algorithms in the literature. A brief introduction to persistent homology is given in Section ~\ref{PHP}. In Section ~\ref{proposed methods}, we describe the proposed method for computing the PHPs and our proposed algorithms for segmenting tumor regions in a WSI. A comprehensive evaluation, comparative analysis of the proposed framework followed by robustness analysis of the proposed methods on an unseen dataset is discussed in Section ~\ref{experiments}. Finally, we conclude with a summary and future directions in Section ~\ref{discussion}.

\section{Related Work}\label{related work}
Existing literature on tumor segmentation in histology images can be broadly classified into two categories: 1) hand-crafted feature based methods and 2) data-driven deep feature based algorithms. In this section, we review previous work regarding the tumor segmentation on images of H\&E stained slides. Literature review on other related methods is covered in relevant sections, where appropriate.

A wide range of studies have been published  on the use of texture, morphological and color features for tumor segmentation. Perception-based
features \cite{bianconi2015discrimination}, local binary patterns (LBP) along with contrast measure features \cite{linder2012identification}, color graphs \cite{altunbay2010color} , Gabor and histogram features \cite{khan2013hymap, kather2016multi}, bags-of-superpixels pyramid \cite{akbar2015tumor}, have been used to segment tumor rich areas. Weakly supervised multiple clustered instance approaches \cite{xu2014weakly, xu2012multiple} have also been proposed for segmentation of tumor in tissue micro arrays (TMAs) of colon cancer images, whereby bags of selected patches are generated to learn the model in a multiple instance framework. The multiple clustered instance model learns from a set of general features like the L*a*b color histogram, LBP, multiwavelet transforms, and scale invariant feature transforms. However, the selection of an optimal set of features for fully or weakly supervised learning is an onerous task and poses the risk of over-emphasizing some particular features of a dataset. Moreover, a major shortcoming of the above algorithms is the fact that their scope is mainly limited to hand-picked visual fields or TMAs. In a clinical setup, a tumor segmentation solution should be capable of scaling the results to the WSI level.  

Deep learning has recently produced exceptional performance on tasks in computer vision \cite{krizhevsky2012imagenet} and in medical image processing \cite{bejnordi2017diagnostic, chen2017dcan, sirinukunwattana2017gland}. One of the well-known methods for segmentation is to learn  a set of hierarchical features by employing a combination of down and up sampling convolution layers, such as U-net  \cite{ronneberger2015u}. These approaches perform reasonably well for pixel level segmentation but are computationally expensive and may encounter the vanishing gradients problem while training.  \cite{cruz2017accurate}  presented a CNN framework for tumor detection in breast histology images. Our CNN architecture for tumor segmentation has some basic similarities with the proposed framework in \cite{cruz2017accurate}. However, our CNN model is relatively deep, enabling the model to learn a set of features at various levels of abstraction. In a supervised learning environment, one can think of deep learning features as a set of data-driven features, learnt by using back propagation (BP). The BP algorithm penalizes the kernel maps by forcing them to learn from their mistakes on the training dataset. Instead of relying on a set of handcrafted features, deep learning models learn the set of optimal features without human intervention. 

Each of the several physical processes involved in creating a WSI, starting with the original biopsy or resection and ending with laying a 3$\si{\micro\metre}$ thick section on a glass slide, is random with respect to orientation. Therefore the textural and geometric features in a WSI will have a random orientation, though the orientations of different features may well
be correlated with each other. However, deep learning models and especially CNNs find difficulty in learning the rotationally invariant charcterstics of an input image \cite{sabour2017dynamic}. In contrast, our biologically interpretable PHPs not only capture the degree of connectedness among nuclei but are also invariant to rotational transformations.



\section{Introduction to Persistent Homology}
\label{PHP}

Persistent homology is an algebraic tool, whereby, given a topological space, certain \textit{algebraic invariants} are computed using the structure of that space. It is a fairly recent concept of homology theory, with a wide range of applications in different domains of data analytics including protein structure \cite{xia2014persistent, cang2017integration}, robotics \cite{bhattacharya2015persistent, pokorny2016topological}, neuroscience \cite{curto2017can}, shape modelling \cite{turner2014persistent}, analyzing brain arteries \cite{bendich2016persistent}, classification of endoscopy images \cite{dunaeva2016classification}, mutational profile and survival analysis \cite{wubie2016clustering},  video surveillance \cite{lamar2016persistent}, time series modelling \cite{otter2017roadmap} and natural language processing \cite{zhu2013persistent}. The concept of persistent homology is relatively new for medical image analysis in general and for histology image analysis in particular. 

Persistent Homology Theory is the study of the homology of a filtered
space, by which we mean a sequence

\begin{equation}
\{ \emptyset = X_{0} \subseteq X_1 \subseteq X_2 \subseteq \dots \subseteq X_k = \textbf{X} \}
\end{equation}
This is referred to as a \textit{filtration} of the topological space $X$. Readers are referred to \cite{edelsbrunner2008persistent, carlsson2009topology, ghrist2008barcodes} for a description of the general theory. 

In our case, we are analyzing 2D grayscale images, and each $X_i$ is the
union of closed pixels in a single image. $X_k = X$ is the entire image. These subspaces are so special, and with such nice properties, that
dramatic simplifications are possible in computing the persistent homology. The only homology groups that are non-zero are in dimensions 0 and 1. We have no need to consider what is often a significant aspect of persistent homology, namely the birth and death of homology classes. It is sufficient for our purposes to consider only the Betti numbers $\beta_0(X_i)$ and $\beta_1(X_i)$ for $1 \leq i < k$. Moreover, these Betti numbers can be computed using basic topological ideas, namely a count of connected components, which is a simple and rapid computational procedure.

To explain how we generate the filtration of a given grayscale image,
we suppose for definiteness that the intensity of each pixel is an integer in the range $[0, 255]$. We then select a sequence of integers $0 = t_0 < t_1 < \dots < t_{k-1} < t_k = 256$. These integers are various threshold levels, at which the image is binarized. We define $X_i$ to be the union of closed pixels p, with intensity $I(p) < t_i$, so that $X_0 = \emptyset$; and $X_k = X$. Each grayscale image gives rise to a single filtration. A careful choice of $t_1, t_2, \dots, t_{k-1}$ balances density in $[0, 255]$ to give all (or nearly all) the information needed from the original grayscale image, against
sparsity to make the algorithm run fast. We compute two relevant homology
groups $H_0(X_i)$ and $H_1(X_i)$ for $1 \leq i < k$ as follows. Instead
of using computationally expensive constructs of simplicial topology, as is normal in the literature using persistent homology \cite{{nakane2015homology, nakane2013simple}}, we compute $H_0(X_i)$ by counting the connected components of $X_i$, giving the Betti number $\beta_0(X_i)$ and $H_1(X_i)$ by counting the components of $X \setminus  X_i$, giving the Betti number $\beta_1(X_i)$. For our purposes, it is therefore sufficient to calculate, for each $i$ with $1 \leq i < k$, these two numbers, giving a total of $2k - 2$ length of feature vector. 

\begin{figure}[t]
\centering
\includegraphics[width=1.0\textwidth]{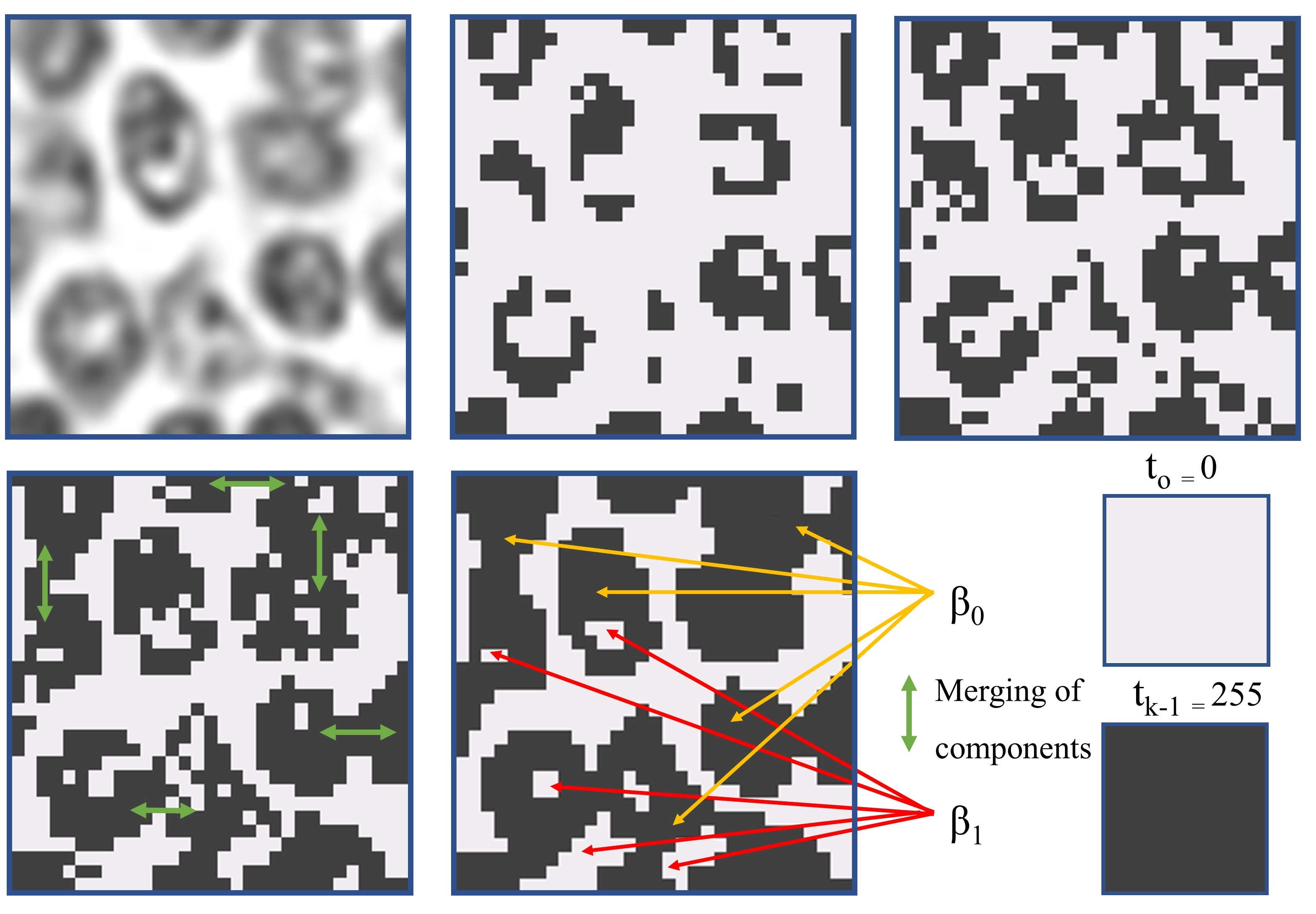}
\caption{An illustration of the filtered space associated to an image. We first show an input image at 40$\times$ magnification, and then four images showing growing sequence of subspaces, obtained by thresholding at increasing values of threshold. $\beta_0$, represents the number of connected components and similarly, $\beta_1$, shows the number of one-dimensional voids (red arrows only show few of them).}
\label{fig:filtration}
\end{figure}

A given small patch as in Figure \ref{fig:filtration} represents a filtered space extracted from a WSI. For a range of $t$ it gives a list of subspaces such that all pixels of previous subspace $X_{i-1}$ are present in $X_i$. To construct PHPs we recorded the rank of homology groups $(H_0(X_i), H_1(X_i))$ for an entire range of  $t$, such as $0 < t_1 < t_2 < \dots < t_{k-1}$. We can also trade a considerable improvement in speed of computation for a negligible loss in information by restricting ourselves to use some well chosen values of $t$. The algorithm for computing PHPs is the main foundation for fast tumor segmentation. In this case, the \textit{algebraic invariants} turn out to be nothing more complicated than whole numbers (ranks of homology group). Hence we do not need to build computationally expensive simplical complexes in order to compute persistent features.  This provides an alternative approach to simplicial homology for 2D images. 

\section{The Proposed Approaches}
\label{proposed methods}
In this section, we present two proposed approaches based on PHPs. We first describe the workflow of the fast tumor segmentation and then explain a variant for accurate tumor segmentation. For a given WSI, we divide it into patches of 256$\times$256 at  20$\times$ magnification. The problem then reduces to classification of each patch as either tumor or non-tumor. 

\subsection{Fast Tumor Segmentation}
The algorithm for fast tumor segmentation is established on three pivotal steps: \textit{1)} an efficient way of computing PHPs, \textit{2)} selection of representative images from the activation maps of a convolutional network, and \textit{3)} a novel algorithm for patch classification.  

\subsubsection{Persistent Homology Profiles}
As mentioned in Section ~\ref{introduction}, tumor nuclei carry atypical characteristics and exhibit chromatin texture allowing us to  distinguish them from non-tumor nuclei. Here we characterize these phenomena with the help of persistent homology for tumor regions in CRC histology images. Our topological features provide a global description of a finite metric space ($I$) by finding the relationships among data points (pixels), contrary to textural and geometrical features where precise distances, angles and spatial arrangement are important. For a given patch $I$, we derive two statistical distributions by recording the ranks of their $0^{th}$ and $1^{st}$ dimensional homology groups, denoted by $\beta_0$ and $\beta_1$. We refer to these statistical distributions as \textit{persistent homology profiles}. 

Histology specimen attains high variability (in colour) mainly due to non-standardization in staining protocols. To overcome this problem we first perform stain deconvolution \cite{ruifrok2001quantification} separating an RGB image patch into three channels, Hematoxylin, Eosin and background. For follow up analysis, we only use the Hematoxylin channel to improve consistency in intensity appearances. To extract the topological features, we binarize the Hematoxylin channel to record the corresponding Betti numbers ($\beta_{0}$, $\beta_{1}$). Rather than relying on a hand-picked threshold value, we observe the topological features for a number of different thresholds $0 = t_0 < t_1 < \dots < t_{k-1} < 256$ as explained above. Later, we convert each statistical curve into a discrete probability distribution, scaling the values so that the area under each curve is one. 

\begin{figure}[t]
\centering
\includegraphics[width=1.0\textwidth]{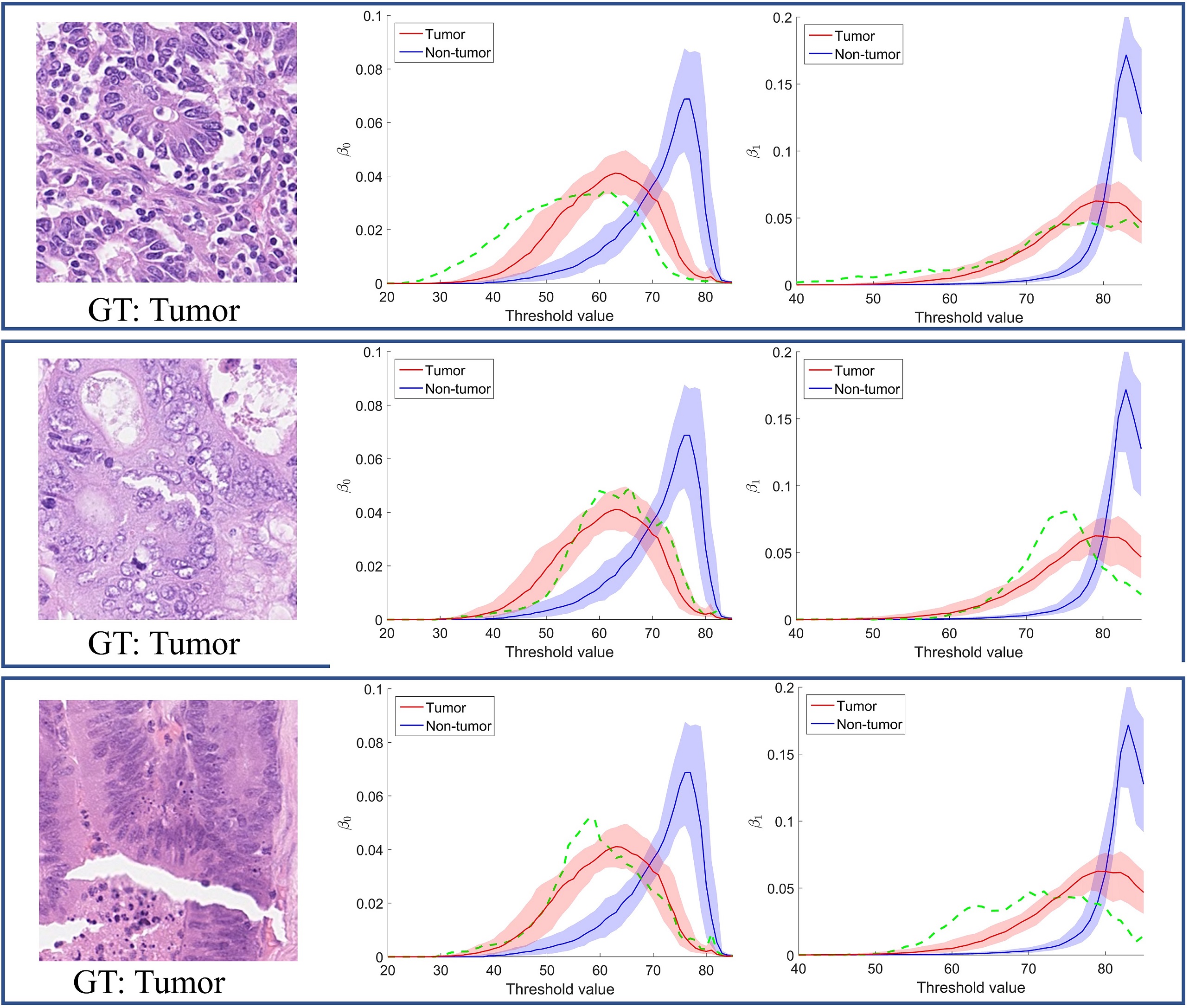}
\caption{ An example of persistent homology profiles (PHP) for the selected tumor patches (left): original images with ground truth (GT), (middle): PHP for $\beta_{0}$, (right): PHP for $\beta_{1}$. The shaded regions in $\beta_{0}$ and $\beta_{1}$ show the first and third quartile of the exemplar patches, whereas the green dotted line shows the PHP of selected patch.}
\label{fig:normal_php}
\end{figure}

\begin{figure}[t]
\centering
\includegraphics[width=1.0\textwidth]{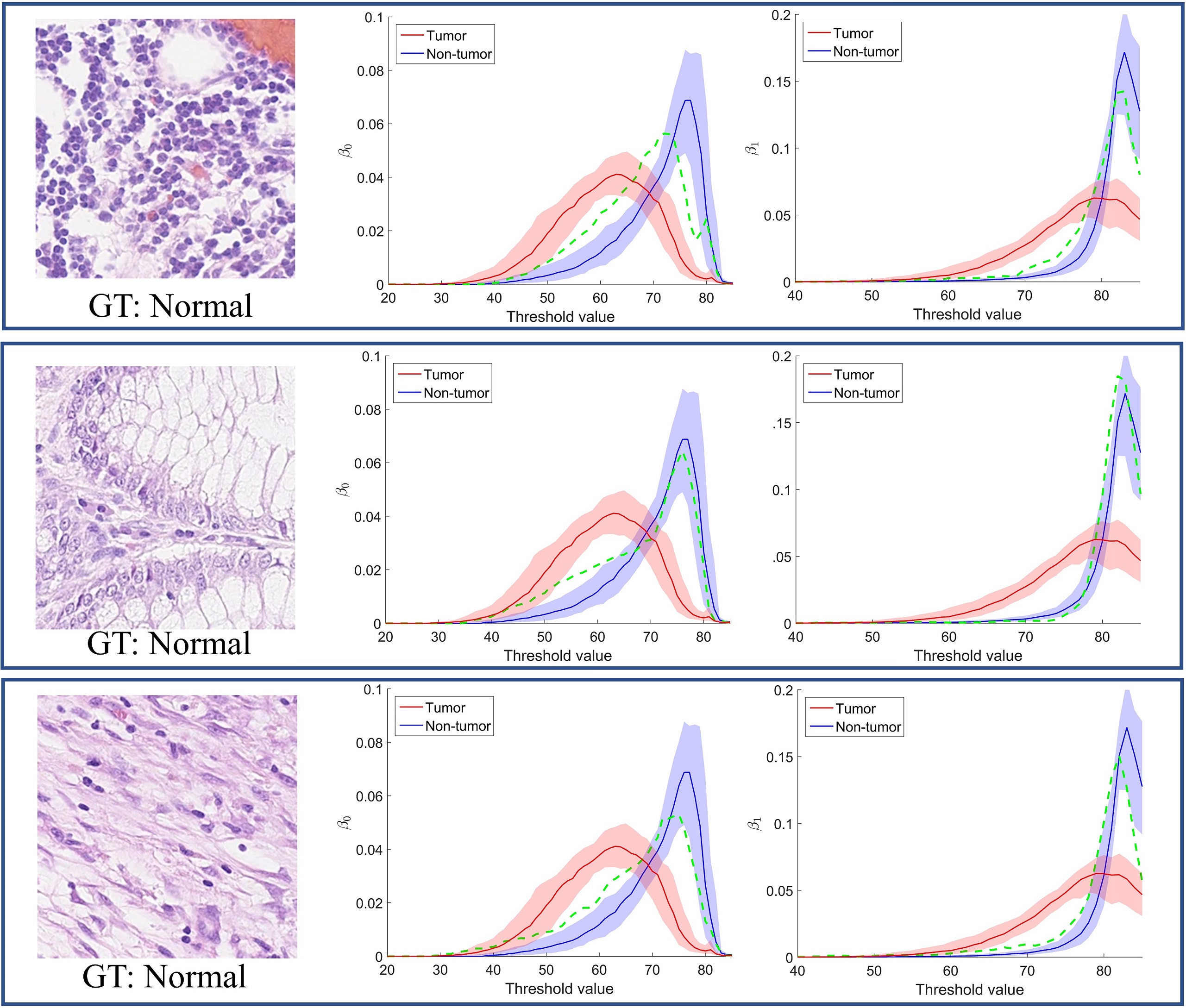}
\caption{ Another example of persistent homology profiles (PHP) for selected non-tumor patches  (left): original images with ground truth (GT), (middle): PHP for $\beta_{0}$, (right) PHP for $\beta_{1}$ . }
\label{fig:tumor_php}
\end{figure}

Differences between tumor and non-tumor regions are reflected in
their homology invariants. This can be seen in Figure \ref{fig:normal_php} and \ref{fig:tumor_php}, which show the curves  representing median of our PHPs for selected exemplar patches from a CNN (explained in Section ~\ref{sec:exemplar_selection} ) for both tumor and normal classes, with first ($Q_1$) and third ($Q_3$) quartile.  The green dotted line shows the PHPs ($\beta_{0}$, $\beta_{1}$) for image patches as shown in the first column. It is worth mentioning here the magnitude of derived PHPs is less relevant instead, the pivotal aspect is the noticeable trend in growth of homology classes. As we start increasing the threshold $t$ from lower limit ($t_0$) to upper limit ($t_{k-1}$), the filtering subspace propagates from an empty set to the entire topological space. Since tumor regions carry more irregularity in their shape, size and tumor nuclei lie relatively close to each other filling the inter-cellular cytoplasmic space, their homology ranks ($\beta_0, \beta_1$) do not show rapid change while merging and forming into new classes as compared to those for non-tumor regions. 

\subsubsection{Selection of Exemplar Patches}
\label{sec:exemplar_selection}
\begin{figure}[t]
\centering
\includegraphics[width=1.0\textwidth]{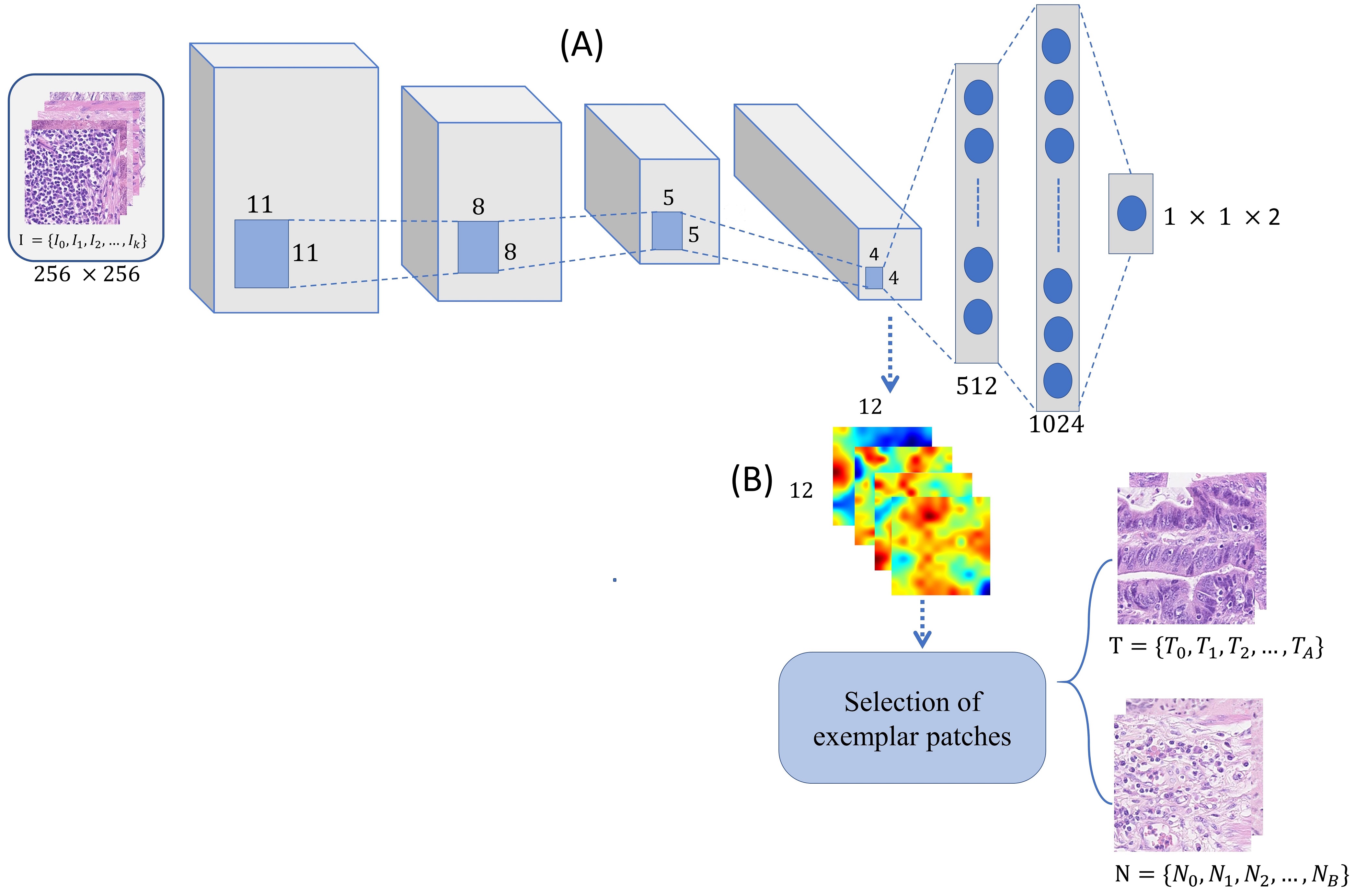}
\caption{ (A) A schematic illustration of our deep convolution neural network (B) An overview of the proposed exemplar selection algorithm based on activation maps of the last convolution layer. 
}
\label{fig:CNN}
\end{figure}

We extract exemplar patches by training a deep CNN. The architecture of CNN is inspired by \cite{krizhevsky2012imagenet} with some modifications, as shown in Figure \ref{fig:CNN}. The objective here is to infer a set of representative patches from the entire training dataset for both tumor and non-tumor classes by exploring the learned activation maps from the last convolution layer. We then compute the PHPs of selected patches in order to measure the value of divergence from an input patch as described in the next section. 

Let us consider a convolution layer with its corresponding activation maps $\alpha \in \mathbb{R}^{W \times H \times Z}$, where $Z$ represents the depth of activation maps of spatial dimension $H \times W$. The activation maps from convolution layers emphasize different tissue parts for different layers. In the first layer, neurons activate for a combination of low-level features like edges on nuclei boundaries and chromatin material found  with in nucleus. The middle layers get more sense of object localization by learning the most discriminitaive regions within tissue patch. The top convolution layer neurons reflect higher level features of tissue components and have high activations around cluster of nuclei with in an input image. The idea here is to deduce the 3D activation maps of the last convolution layer to a scalar value that can be used as an indication of the significance of each patch with respect to the activation maps. We first define a mapping function closely related to \cite{zagoruyko2016paying} to flatten the 3D activation maps to 2D across the $Z$ dimension as below,
\begin{equation}
F(w,h) = \sum_{z=1}^{Z} \left|\alpha_{(z)}^{w,h}\right| ^2‎‎,
\end{equation}
Here, the mapping function gives more weight to neurons with high activations. It is worth noting that normalizing the flatten 2D activation map is important for follow-up analysis. We compute the median value of $F(w,h)$  to find the central tendency of the 2D activation map, that later assist in exclusion of unimportant patches in the training dataset. Similarly, for the entire training dataset we get an $M \times 1$ vector by computing the $median(F(w,h))$ for each patch, where $M$ represents the number of patches.

The last step is to find a set of exemplar patches that can later be used as representatives for the entire training dataset. One way of choosing the exemplar patches is to find the most highly activated patches from the $M \times 1$ matrix. The only pitfall of following such a method is that we may end up selecting patches representing a certain type of tumor or normal tissue. In order to avoid this scenario, we compute the interquartile range (IQR) of $M \times 1$ matrix separately for the tumor and non-tumor classes. Furthermore, we split the IQR of each class into $Q$ same size bins and select the value that lies closest to the median of corresponding bin, where $Q$ represents the number of exemplar patches for each class. This method for selection of exemplar patches differs from \cite{qaiser2016persistent} and is capable of handling a much larger dataset as in Section ~\ref{sec:results exemplars}.   

\subsubsection{Patch Classification}

For patch classification, we first transform the PHPs into discrete probability distributions. We then compute the symmetric KLD to measure the distance between the PHPs of an input patch $I$ and the PHPs of an exemplar patch $E$ as defined below,
\begin{equation}
D_{KL}(I \parallel E) = \sum_{i}  I(i) \log\frac{I (i)}{E(i)}‎‎,
\end{equation}
where $I$ represents the \textit{true} representation of data and $E$ represents an \textit{approximation} of $I$. The symmetrised, non-negative KLD is defined as 
\begin{equation}
d_{I,E} =  D_{KL}(I \parallel E) + ‎‎D_{KL}(E \parallel I).
\end{equation}

We compute a vector of distance values to exemplar patches  $D(I) = ( d_{I,T_1},$ \\$d_{I,T_2},...,d_{I_A,T_A},$  
$d_{I,N_1},d_{I,N_2},...,d_{I,N_B})$, containing divergence values for input patch PHP profiles of I and all exemplar tumor $T = \{T_1,...,T_A\}$ and non-tumor exemplar patches $N = \{N_1,...,N_B\}$. We derive a similarity measure from the KLD values computed in (5) and compare the total similarity scores to the $k$ nearest tumor and non-tumor patches as follows:
\begin{equation}
\sum_{j=0}^{k_{t}} e^{-d_{I,T_j}c}  > \sum_{j=0}^{k_{n}}  e^{-d_{I,N_j}c},
\end{equation}
where $c$ is a constant in the interval $(0, \max(d))$ and $k_{t}$, $k_{n}$ denotes the number of nearest tumor and non-tumor patches according to (5), and $k= k_t +  k_n$.

\subsection{Accurate Tumor Segmentation}
In this variant of our algorithm, we combine deep convolutional and persistent homology features. For this we trained a CNN as shown in Figure \ref{fig:CNN} to extract features from the last fully connected layer. We then fed the extracted features to a Random Forest (RF) regression model separately for the topological and deep convolutional features. Finally we propose a multi-stage ensemble strategy to combine the two RF regression models. The main objective of this method is to combine \textit{best of the both worlds}. The key contribution of the topological features is to capture the underlying connectivity whereas the CNN tends to learn the data driven features.

\subsubsection{Deep Convolutional Features}
The CNN architecture contains four convolutional layers followed by an activation function and a max-pooling operation. Additionally, it contains two fully connected layers and the softmax classification layer to predict the label of each patch as tumor or non-tumor. Instead of using a rectified linear unit ReLU as activation function, we use an exponential linear unit (ELU) \cite{clevert2015fast}, as it enables faster convergence and also reduces the vanishing gradient problem. A dropout layer at the end of the second fully connected layer is placed to overcome the overfitting problem. The CNN was trained to minimize the overall cross entropy loss $L$ ~\ref{eq:cross_entropy}. 
\begin{equation}
L(g, y) =  - \sum_{x} g(x)log ( y(x))
\label{eq:cross_entropy}
\end{equation} 
\\
where for input $x$, $g$ represents the ground-truth label (0 for tumor and 1 for non-normal) and $y$  is the probability of the tumor predicted by the CNN. The fully connected layers contain non-linear combinations of learned features from the convolution layers. We extracted CNN features for the training dataset after the last fully connected layer just before the softmax classification layer. For each patch of the training dataset we obtained a feature vector of size $(1 \times 1 \times 1024)$. 

\subsubsection{Ensemble Strategy}
After obtaining topological and convolutional features, we concatenate both PHPs ($\beta 0, \beta 1$)  to form a combined topological feature vector. We then train the RF regression model separately for both types of feature. We optimize the RF model with an ensemble of 200 bagged trees, randomly selecting one third of the variables for each decision split and setting the minimum leaf size to 5.

We combine the probability of both regression models ($O_1, O_2$) as in (7), where $O_1$ represents a regression model of topological features and similarly $O_2$ represents a regression model of convolutional features. The multi-stage ensemble strategy  follows  two  alternative routes: a) averaging the outcome probabilities of $O_1$ and $O_2$ to predict the output label where both regression models agree b) for the remaining few patches ($\approx 1\% $ from the test data) where the average probabilities lies in range $0.49 - 0.51 $. We refer to these as \textit{critical patches} and we assign the output label for those patches by rounding the probabilities from $O_1(x)$ as in our experiments based on two dataset we observed the comparatively high F1-score with the setting.

\begin{equation}
  \hat O(x)=\begin{cases}
    0, & \text{if $ \frac{(O_1  + O_2)}{2} < 0.49 $}.\\
    1, & \text{else if $ \frac{(O_1  + O_2)}{2} > 0.51 $}. \\
    \lfloor O_1(x) \rceil	& \text{otherwise (rounding)}
    
  \end{cases}
\end{equation}

\section{Experiments and Results}
\label{experiments}

\subsection{Dataset and Experimental Setup}

\paragraph{The Warwick-UHCW Dataset} This dataset consists of 75 WSIs of H\&E stained colorectal adenocarcinoma tissues. At the highest resolution, each WSI normally contains more than 10$^{10}$  pixels. The WSIs were digitally scanned at a pixel resolution of 0.275$\si{\micro \meter}$/pixel (40$\times$) using an Omnyx VL120 scanner. The ground-truth for tumor regions were handmarked by an expert pathologist.  For each WSI, we randomly selected 1,500 patches including 750 from non-tumor and 750 patches from tumor regions. In total, we extracted, 75,000 patches for training from 50 WSIs and 37,500 patches for testing from 25 WSIs. The collected dataset for this study is roughly 20 times more than that in \cite{qaiser2016persistent} and at least 2 times more than in \cite{qaiser2017tumor}. For generating the tumor probability map of a WSI, we first split the given WSI into patches and then applied our methods to each patch. 

\paragraph{The Warwick-Osaka Dataset} This dataset contains 50 H\&E stained histology WSIs of colorectal tissue. The WSIs were scanned at a pixel resolution of 0.23$\si{\micro \meter}$/pixel (40$\times$) using a Hamamatsu NanoZoomer 2.0-HT scanner. The ground-truth for this dataset were handmarked by two expert pathologists and the cases were identified as  belonging to 6 categories, including 11 cases of adenoma, 14 moderately differentiated, 6 poorly differentiated, 10 well-differentiated, 8 healthy and 1 signet case. The inclusion of normal cases in this dataset helps in evaluating the robustness  of the proposed methods as discussed in Section ~\ref{sec:results healthy}. Similarly to the Warwick-UHCW dataset we randomly selected 1,500 patches (750 tumor, 750 non-tumor) from each WSI.

\paragraph{Experimental Setting} For our proposed methodologies we split a WSI into manageable patches of $256\times256$ for training as well as testing. In order to counter overfitting we performed data augmentation by rotating $(0^{\circ}, 90^{\circ}, 180^{\circ}, 270^{\circ})$, flipping (horizontal or vertical axis), and perturbing the color distribution (hue variation) of both the training datasets. The weights for a CNN were initialized by using Xavier initialization as in (8).
\begin{equation}
	l = \sqrt{ \frac{3}{ (N_{in} + N_{out})}}
\end{equation}
where $N_{in}$ and $N_{out}$ represents the number of input and output neurons, respectively. The selected weight initialization approach tends to restrict the magnitude of gradients from excessive shrinking or growing during the training process. The network learns the weights by using the mini-batch gradient descent algorithm by selecting a batch size of $100$. During the training phase, the initial learning rate was set to $0.0001$ and an Adam optimizer was employed instead of conventional gradient descent algorithm. In addition, a dropout layer (dropout rate 0.5)  was placed between the two fully connected layers to overcome the interdependence among intermediate neurons and to increase the robustness of the trained network. For the fast tumor segmentation, we separately chose 128 exemplar patches for both tumor and non tumor, $c=0.2$ for similarity measures and   $k=11$ for $k$-NN.

\paragraph{Evaluation} We compute the F1-score to evaluate the performance of different approaches as a harmonic mean of precision and recall as defined below,
\begin{equation}
F_1 = 2 \times \frac{P_r \times R_e }{(P_r + R_e)}; \qquad P_r =  \frac{T_p }{T_p +  F_p};
\qquad R_e =  \frac{T_p }{T_p +  F_n}
\end{equation} 
where $T_p$, $F_p$, and $F_n$ represents the number of true positive, false positive, and false negative patches. For a  given test dataset, the correctly identified patches are classified as either true positives or true negative, misclassified predictions are categorized as false positives, and true negatives.

\begin{figure}
\centering
\includegraphics[width=1.0\textwidth]{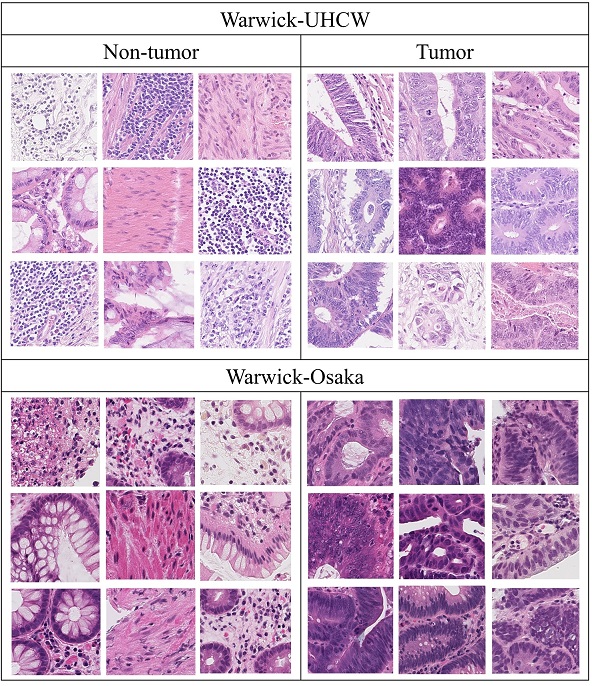}
\caption{Representative tumor and non-tumor  patches from the Warwick-UHCW and  the  Warwick-Osaka datasets. }
\label{fig:exemplars}
\end{figure}

\subsection{Comparative Analysis}
\subsubsection{Selection of Exemplar Patches}
\label{sec:results exemplars}
One of the most critical parts in the fast tumor segmentation is the selection of exemplar patches. Hence the objective of this experiment is to investigate a few options for selection of exemplar patches. The experiment is conducted on 75 colorectal  adenocarcinoma WSIs from the Warwick-UHCW dataset. The following algorithms were selected for comparison and as an alternative to the proposed method of exemplar selection. 
We start with a random selection of exemplar patches from the training dataset for both tumor and non-tumor classes. We repeat this process 10 times before concluding the final results and the reported results (Table \ref{tab:exemplars}) show the mean precision, recall and F1-score. In $k$-means clustering the training dataset was partitioned into $k$ clusters with respect to their RGB intensities. Then we selected those patches that lie closest to the cluster centroids as exemplar patches, one patch per centroid. The third algorithm for comparison is to select highly activated patches as exemplars from a CNN  as proposed in \cite{qaiser2016persistent}. For a fair comparison, an equal number of exemplar patches were selected for each of the above mentioned algorithms.

\begin{table}
\begin{center}
  \caption{Comparison of various options for selecting the exemplar patches; reults on the Warwick-UHCW dataset.}

  \begin{tabular}{ |p{6.5cm}|p{1.5cm}|p{1.5cm}| p{1.5cm}|  }

\hline
Method & Precision & Recall & F1-score\\
\hline
CNN  (Section ~\ref{sec:exemplar_selection}) & {0.9272} & \textbf{0.8513} & \textbf{0.8999} \\
\hline
CNN (highly activated) \cite{qaiser2016persistent} & 0.9112 & 0.8309 &	0.8692\\
\hline
$k$-means &\textbf{0.968} & 0.7559 & 0.8489 \\
\hline
Random selection & 0.8812 & 0.8309 & 0.8553 \\

\hline
\end{tabular} 

 \label{tab:exemplars}
 \end{center}
\end{table}

Table \ref{tab:exemplars} reports the patch based tumor segmentation results for different approaches. Overall the results are in favor of the proposed method. Figure \ref{fig:exemplars} shows a sample of 9 representative patches for tumor and non-tumor class from both the dataset, by using the proposed method. The CNN (highly activated) \cite{qaiser2016persistent} seems a straightforward approach that selects only patches where CNN neurons produce high activations. However for a relatively large dataset the exemplars may be overemphasized by a particular atypical WSI, where we normally have thousands of patches from a single case. Thus, this kind of approach is more suitable for a small dataset. Another downside is inclusion of outliers as exemplars which can easily happen due to lack of precisely marked ground-truth. This argument remains valid for $k$-means and random selection approaches.  

\subsubsection{Tumor Segmentation on Adenocarcinoma Cases}
In this experiment, we evaluate the performance on the UHCW-Warwick dataset of the proposed algorithms in comparison to  some recently published algorithms for tumor  segmentation. The experiment is conducted on all 75 adenocarcinoma WSIs where we used 75,000 randomly selected patches from 50 cases for training and the remaining 37,500 patches from 25 WSIs for testing. For a fair comparison, we retrained the selected algorithms on our dataset except for HyMap \cite{khan2013hymap}, which is an unsupervised method. For comparative analysis we selected only algorithms that are closely related to CRC image analysis provided that their source codes were released with proper implementation details and comments. The features of selected algorithms are as follows: 

\begin{figure}
\centering
\includegraphics[width=1.0\textwidth]{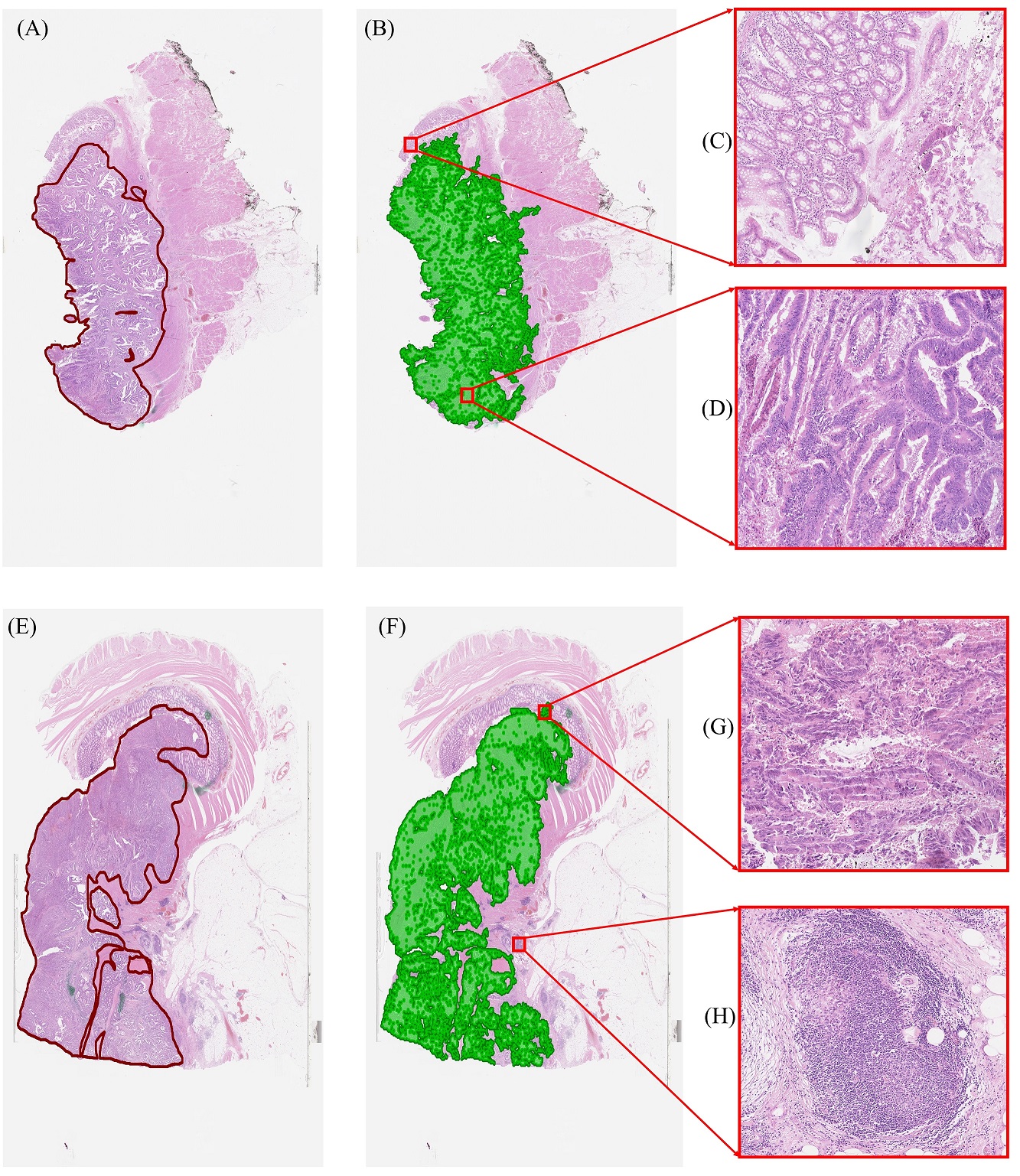}
\caption{Results of accurate tumor segmentation on the whole-slide image (WSI) level. (A)\&(E) input WSIs with annotated ground-truth, (B)\&(F) tumor segmentation results, green region showing the predicted tumor areas, (C)\&(H) zoom in regions containing true negatives (D)\&(G) showing a sample of the true positive tumor segmented regions.}
\label{fig:wsi_uhcw}
\end{figure}

\begin{itemize}
  \item Multi-class texture analysis \cite{kather2016multi} or MCTA: This method first computes a set of textural features on a given patch including lower-order and higher-order histogram statistics, local binary patterns, gray-level co-occurrence matrix, Gabor filter, preception-like features and then fed these features into radial-basis function (RBF) support vector machine (SVM).  
  \item HyMap \cite{khan2013hymap}: This is an unsupervised algorithm for tumor segmentation that classifies each pixel as hypo or hyper cellular by computing Gabor filter, texture energy and phase-gradient features followed by ensembling the projections for each feature. Since this is an unsupervised method, we did not retrain it for our experiments.
  
  \item ConvNet CNN$_{3}$ \cite{cruz2017accurate}: This method poses the cancer detection as a two-class problem by assigning an `invasive or non-invasive' label to each patch. The framework is a combination of convolutional layers followed by pooling operations and fully connected layers. The reported results suggest the CNN$_{3}$ outperformed the other counterparts on breast tissue so we used only CNN$_{3}$ for our experiment.  
  
  \item Texture-based tumour viability image analysis \cite{turkki2015assessment} or TVIA: It classify a given patch as viable or non-viable or as `other tissue parts'. They computed local binary patterns and local contrast measures at the patch level and fed the extracted features into a SVM model. This  algorithm is not directly related to tumor segmentation. We therefore evaluate it only as a two-class problem, that is, as a decision whether to classify a patch as tumor  or non-tumor. 
\end{itemize}

\begin{table}
  \centering
  \caption{Tumor segmentation results on the Warwick-UHCW dataset}
  \begin{tabular}{ |p{6.9cm}|p{1.3cm}|p{1.3cm}| p{1.3cm}|  }

\hline
Method & Precision & Recall & F1-score\\
\hline
Fast Tumor Segmentation (PHP) & \textbf{0.9272} & 0.8513 & 0.8999  \\
\hline
Accurate Tumor Segmentation (PHP+CNN) & 0.9267 & \textbf{0.922} & \textbf{0.9243} \\
\hline
HyMap \cite{khan2013hymap} & 0.6851 & 0.8600 & 0.7626 \\
\hline
ConvNet CNN $_3$  \cite{cruz2017accurate} & 0.856 & 0.867 & 0.8615 \\
\hline
MCTA \cite{kather2016multi} & 0.8701 & 0.8834 & 0.8767 \\
\hline
TVIA \cite{turkki2015assessment} & 0.8224 & 0.8641 & 0.8427 \\
\hline
\end{tabular} 
 \label{tab:uhcw}
\end{table}

Table \ref{tab:uhcw} reports comparative results from the experiment and Figure \ref{fig:wsi_uhcw} shows qualitative results for the accurate tumor segmentation on the WSI level. Our PHP+CNN based accurate tumor segmentation produces the best results in terms of recall and F1-score, whereas the PHP based fast tumor segmentation performs best for the precision metric. The PHP+CNN outperformed the other competing methods by a reasonable margin. It is encouraging that incorporating deep features along with topological features boosts the overall performance. Generally CNN models struggle to capture the rotation or viewpoint invariance of learned object. To overcome this deficiency we need to perform flipping or several arbitrary rotations on our images in form of augmentation to make our model more generalize. Data augmentation overcomes the rotational invariance to some extent but it is still a non-trivial task to produce all possible rotations while training a CNN classifier. In contrast, PHP captures the rotational invariance by emphasizing the merging and forming of homology classes so no matter how much a patch is rotated the PHP will remain persistent. The PHP also captures the biological phenomenon that the    connectivity among tumor and non-tumor nuclei is significantly different. As compared to CNNs, the PHP based fast tumor segmentation algorithm only consults a number of exemplar patches from both classes in predicting the outcome of a patch. In our previous work \cite{qaiser2016persistent}, we observed that a similar approach performs marginally better than CNN. This offers a trade-of between accuracy and efficiency with reliable outcomes.

\subsubsection{Tumor Segmentation on Adenoma, Carcinoma and Healthy Cases }
The goal of this experiment is to test the generalization of the proposed algorithms on another dataset that consists of different types of epithelial tumors and healthy cases.  The experiment is conducted on 50 WSIs of the Warwick-Osaka dataset. We perform 2-fold cross-validation by selecting half of the dataset for training and the remaining half for testing. We then perform the same experiment by switching the training and test datasets. Similarly to the experiment with the Warwick-UHCW dataset we perform comparative analysis on the aforementioned selected algorithms and by retraining them on the Warwick-Osaka dataset. 

\begin{table}
  \centering
  \caption{Tumor segmentation results on the Warwick-Osaka dataset}
  \begin{tabular}{ |p{6.9cm}|p{1.3cm}|p{1.3cm}| p{1.3cm}|  }

\hline

Method & Precision & Recall & F1-score\\
\hline

Fast Tumor Segmentation (PHP) & 0.8259 & 0.8019 & 0.8137 \\
\hline
Accurate Tumor Segmentation (PHP+CNN) & \textbf{0.8311} & \textbf{0.8235} & \textbf{0.8273} \\
\hline
HyMap \cite{khan2013hymap} & 0.6469 & 0.7228 & 0.6827 \\
\hline
ConvNet CNN$_3$ \cite{cruz2017accurate} & 0.6927 & 0.8446 & 0.7612 \\
\hline
MCTA \cite{kather2016multi} &  0.7050 & 0.7419 & 0.7229 \\
\hline
TVIA \cite{turkki2015assessment} &  0.6993 & 0.7240 & 0.7114 \\
\hline
\end{tabular} 
 \label{tab:osaka}
\end{table}

Table \ref{tab:osaka} shows the results from 2-fold cross-validation. It can be observed that the proposed methods for all 3 metrics perform well on an independent dataset. One of the arguments for ascribing relatively low performance by MCTA, HyMap and TVIA is the selection of textural features, especially local binary patterns, local contrast measures, histogram statistics and the gray-level co-occurrence matrix, that are sensitive to stain variations and image blurring. An additional cause for the poor performance of these programs is that, due to varying staining protocols at different centers, the morphological appearance of lymphocytes and benign epithelial nuclei from one
center can resemble that of malignant epithelial nuclei from another.

\subsubsection{Results for Healthy Cases}
\label{sec:results healthy}
\begin{figure}[t]
\centering
\includegraphics[width=1.0\textwidth]{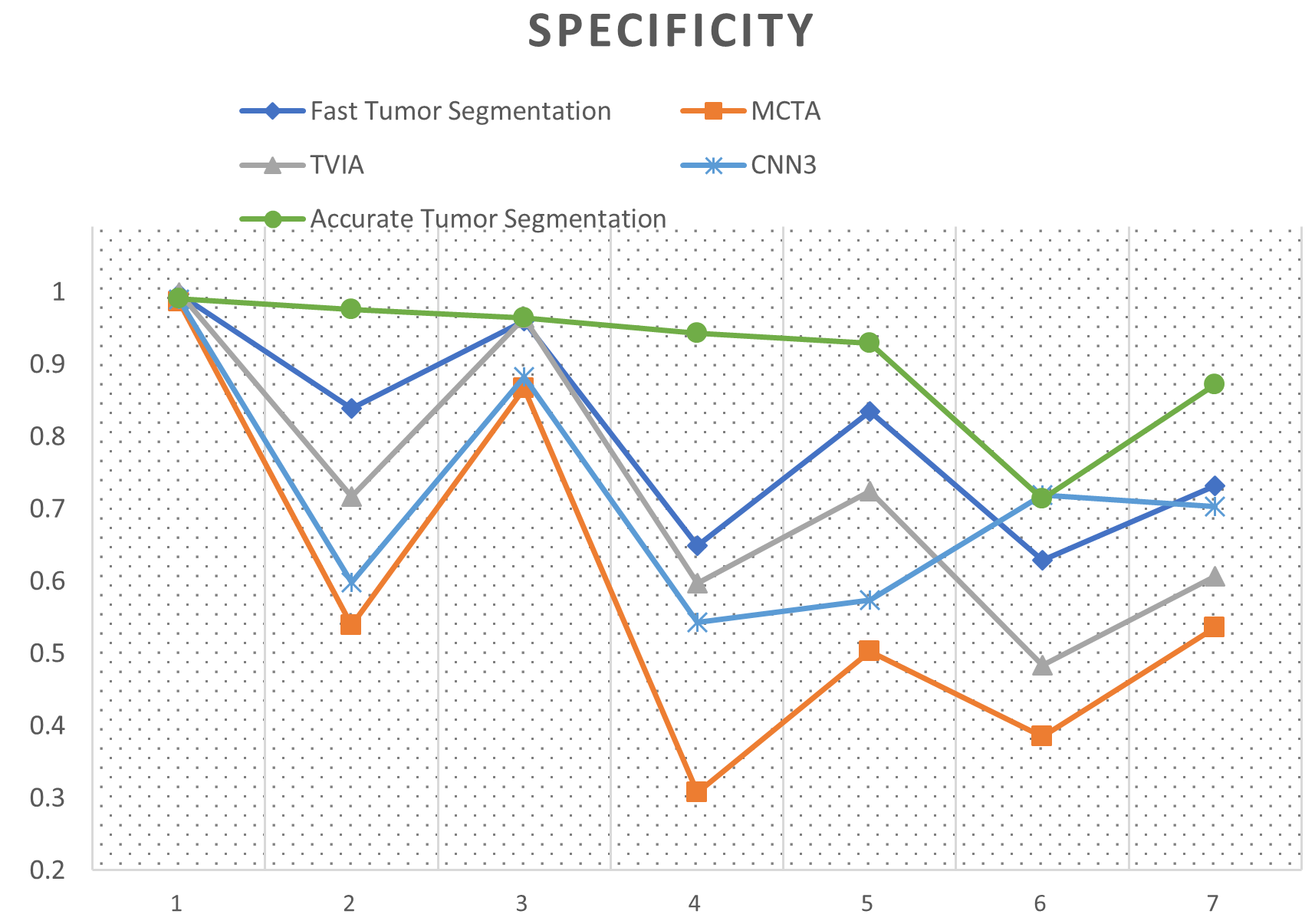}
\caption{ Specificity curve for tumor segmentation algorithms on healthy cases from the Warwick-Osaka dataset. The horizontal  axis shows the number of healthy cases selected, and the vertical axis shows the specificity.}
\label{fig:healthy}
\end{figure}

This experiment compares the performance of different tumor segmentation algorithms for healthy cases in the Warwick-Osaka dataset. This is important for routine clinical practice as well as in cancer screening studies, which involve a large number of normal cases. The most challenging sections are healthy epithelial and lymphocyte regions where nuclei are densely packed and pose difficulties in identifying those regions as non-tumor. We evaluate the performance of this experiment by the specificity metric (also known as the true negative rate), that measures the involvement of negative samples misclassified as positive. Figure \ref{fig:healthy} shows the summarized results for all the healthy cases involved in this study. The specificity analysis demonstrates the effectiveness of the proposed ensemble strategy for accurate tumor segmentation, showing that a model relying on agreement between topological and deep features outperforms all competing approaches.

\subsubsection{Robustness Analysis}
The aim of performing this experiment is to evaluate the robustness of the proposed tumor segmentation algorithms by training on the Warwick-UHCW dataset and testing on the Warwick-Osaka dataset which contains cancerous and healthy cases. One of the most challenging aspects of processing H\&E WSIs is to overcome the non-standardized parameters involved in slide preparations like tissue sectioning, staining duration, dyes and formalin concentration \cite{veta2015assessment}. In order to become a part of the routine diagnosis an automated tumor segmentation algorithm should be resilient to these data variations. With that in mind we perform this experiment by  keeping the parameters of the algorithms fixed during the training and testing.

\begin{figure}[t]
\centering
\includegraphics[width=1.0\textwidth]{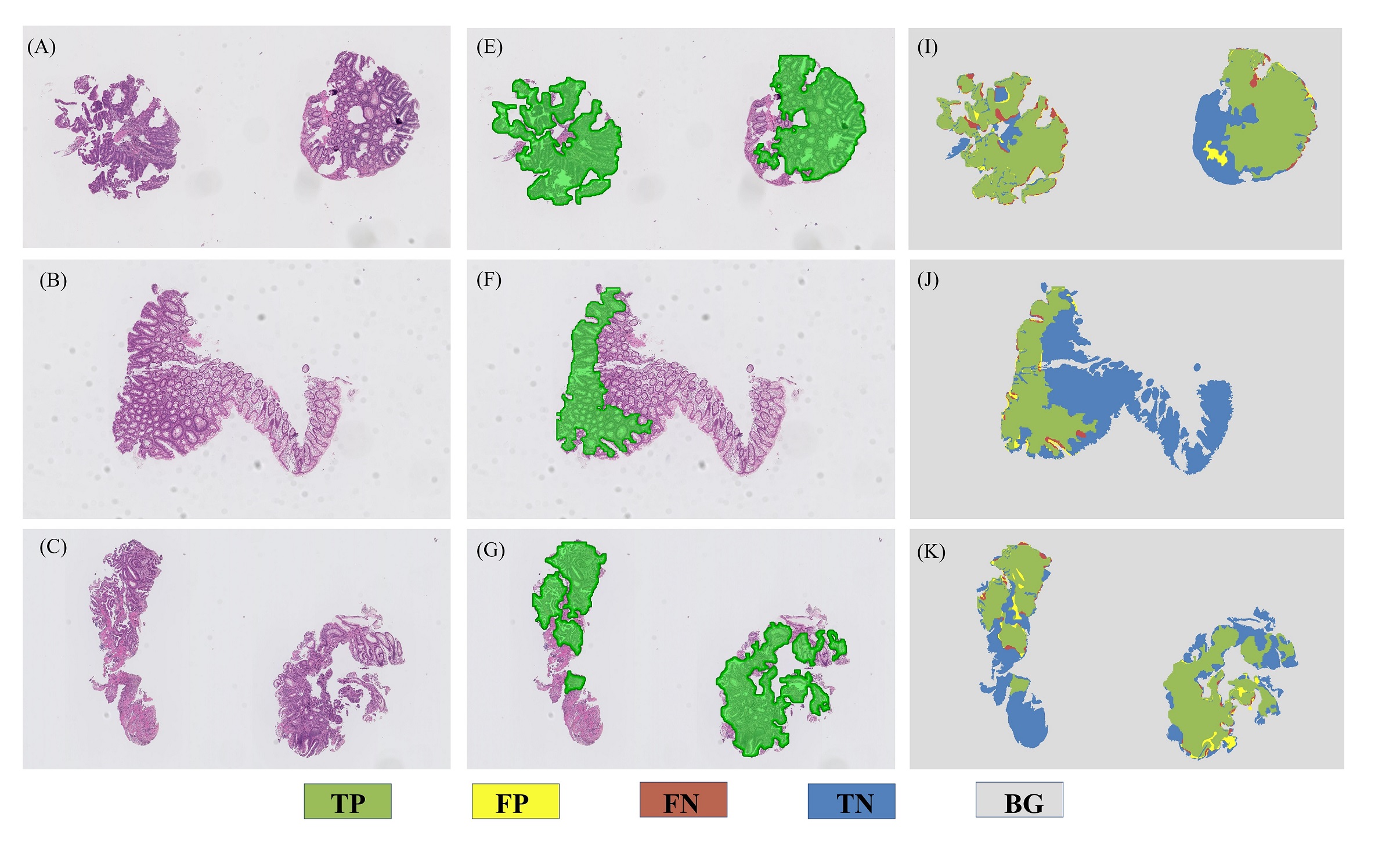}
\caption{Robustness Analysis: results for the fast tumor segmentation on selected whole-slide images (WSI) from the Warwick-Osaka dataset. (A)-(C) input WSIs, (E)-(G) predicted tumor regions, (I)-(J) results for true positives (TP), false positive (FP), false negative (FN), and true negative (TN) regions.}
\label{fig:wsi_osaka}
\end{figure}

Table \ref{tab:robust} reports the results for robustness analysis and Figure \ref{fig:wsi_osaka} shows some qualitative results. The proposed methods attain the highest accuracy among different methods, demonstrating their robustness on strongly cross-validated data. The stain variability  in both the datasets can be observed in Figure \ref{fig:exemplars}, which shows the selected exemplars from both the  datasets. Regardless of stain variations in both the datasets, the degree of connectivity among tumor and non-tumor class is notably distinguishable and that leads to better performance of the proposed methods. It is interesting to note that the PHP based fast tumor segmentation marginally outperformed the PHP+CNN based accurate tumor segmentation (0.3$\%$). This bodes well for the potential generalizability of fast tumor segmentation. Relatively smaller values of F1 measure for other competing algorithms can also be noticed in Table \ref{tab:robust}. Excluding HyMap, an obvious indication for considerable drop in F1-score is due to the precision measure. On the one hand, the competing algorithms perform well in predicting tumor patches but at the cost of a large number of false positives. In contrast, regardless of stain variations in the two datasets, the fast tumor segmentation algorithm maintains a balance between precision and recall.

\begin{table}[h]
  \centering
  \caption{Results for robustness analysis of different tumor segmentation algorithms.}
  \begin{tabular}{ |p{6.9cm}|p{1.3cm}|p{1.3cm}| p{1.3cm}|  }

\hline
Method & Precision & Recall & F1-score\\
\hline
Fast Tumor Segmentation (PHP)
 & \textbf{0.772} & 0.7890 & \textbf{0.7804} \\
\hline
Accurate Tumor Segmentation (PHP+CNN) & 0.7413 & 0.8172 & 0.7774 \\
\hline
HyMap \cite{khan2013hymap} & 0.6469 & 0.7228 & 0.6827 \\
\hline
ConvNet CNN$_3$ \cite{cruz2017accurate} &0.6234 &  0.8167 & 0.7071 \\
\hline
MCTA \cite{kather2016multi}  & 0.5429 & 0.9586 & 0.6932 \\
\hline
TVIA \cite{turkki2015assessment} & 0.5334 & \textbf{0.9747} & 0.6895 \\

\hline
\end{tabular} 
 \label{tab:robust}
\end{table}

\subsubsection{Runtime Analysis}
This section contains the runtime analysis of different tumor segmentation algorithms. Digitized WSIs are giga-pixel images, so fast tumor segmentation algorithms could play a crucial role in delivering 'real world' diagnostics. For all the algorithms, runtime analysis at the test stage is performed on a static machine having an 8-core processor with 3.1 GHz clock speed, 128 MB of memory and a GTX 1080 Ti graphical processing unit. 

\begin{table}[t]
  \centering
  \caption{Runtime analysis in milliseconds for different tumor segmentation algorithms}
  \begin{tabular}{ |p{7cm}|p{1.7cm}|p{3cm}| p{3cm}|  }

\hline
Method & Time \\
\hline
Fast Tumor Segmentation (PHP) & \textbf{28.8} ms \\
\hline
Accurate Tumor Segmentation (PHP+CNN) & 151 ms  \\
\hline
HyMap \cite{khan2013hymap} &  931 ms  \\
\hline
ConvNet CNN$_3$ \cite{cruz2017accurate} & 97.1 ms \\
\hline
MCTA \cite{kather2016multi}  &  228.12 ms \\
\hline
TVIA \cite{turkki2015assessment}  & 109 ms  \\
\hline
\end{tabular} 
 \label{tab:time}
\end{table}

Table \ref{tab:time} reports the processing time for different algorithms on a patch of size 256$\times$256$\times$3 at 20$\times$. The fast tumor segmentation is less computationally complex and an order of magnitude faster than competing algorithms, specifically $\approx4.2$ times faster than the CNN and $\approx5.2$ times faster than the accurate tumor segmentation. The algorithm for computing the PHPs is the foundation of fast tumor segmentation. The topological features ($\beta_0, \beta_1$) are computed by enumerating the connected components for a given filtered space. Consequently, the fast tumor segmentation algorithm is far less computationally complex than a multi-layer convolutional network. 

\section{Discussion and Conclusions}
\label{discussion}
Visual examination of tissue slides under the microscope to analyze the morphological features is the 'gold standard' for cancer diagnosis \cite{gurcan2009histopathological}. This study aimed at improving the diagnostic workflow by introducing a novel automated tumor segmentation framework for colorectal cancer  histology WSIs. Experimental results conducted on fairly challenging datasets collected from two independent pathology centers demonstrate the efficacy and generalizability of persistent homology in histopahtological image analysis. In this work we present novel topological signatures (PHPs) that, in some respects, resemble clinicians' approach for identification of tumor enrich areas. It is evident from the performance that incorporating topological features and deep convolution features can  enhance the overall performance of a CNN for the task of accurate tumor segmentation. The proposed framework was shown to work well on colorectal epithelial tumors of different histology grades.  

Histological assessments are generally estimated  visually and producing   subjective measures for quantification of morphological features \cite{webster2011quantifying}. Computer-assisted image analysis requires precise annotations at high resolution to effectively train an underlying model. The inevitable fact is that the domain experts are generally not available for this laborious task of providing precise ground-truth at high resolutions. In such circumstances the performance of a trained model may be affected by expert annotation that is carried out too rapidly or without close attention to detail, or by inexpert annotation. After careful consideration, we decided that this could be a relevant factor in both datasets, resulting in a difference in performance of the tumor segmentation algorithms (Table \ref{tab:uhcw} \& \ref{tab:osaka}), regardless of stain and morphological variability. Nearly all the tumor segmentation algorithms experience difficulties in some of the benign epithelial and lymphocytic regions especially where the intra-cellular region displays morphology that is similar to cancer-distorted nucleoplasm. 

WSI scanners are becoming more viable for routine analysis \cite{snead2016validation}, capable of producing hundreds of terabytes of data daily \cite{cooper2012digital, kamel2011trends}. The proposed framework offers a decent trade-off between speed and accuracy. With this change of paradigm, the fast tumor segmentation has enormous potential to overcome this ongoing challenge, reducing subjectivity and the pathologists' workload. It can be observed from the experimental results that careful selection of the exemplar patches can nearly obviate the need for retraining of our algorithm on a new dataset as shown in Table \ref{tab:robust}. One limitation of this work is that parameters  like  the number of exemplar patches and $k$ (as in $k$-means) are empirically selected for this study and may need proper fine tuning, depending upon the data. The selection of a representative subset from a dataset is an active area of research and has several applications in computer vision and natural language processing.  The proposed method for selection of exemplar patches is presented as an application to deep convolution networks. In the literature, there exist dissimilarity based subset selection methods \cite{elhamifar2016dissimilarity, elhamifar2012finding}, but selection of a dissimilarity measure is a non-trival task and computing a dissimilarity matrix for a
large dataset is computationally expensive. Other approaches like loopy belief propagation \cite{murphy1999loopy} can handle a large dataset but does not offer any guarantee of  convergence. In contrast, the proposed fast tumor segmentation method exploits the learned activation maps to deduce the representative patches and is less computationally expensive and more robust to outliers. It is also evident from the experimental results that the accurate tumor segmentation algorithm presents a simple yet meaningful way of combining our novel topological signatures with deep convolution features. An interesting direction could be to treat the homology profiles as temporal data and to explore such temporal information with recurrent networks. In this paper, we restrict ourselves to convolution networks
which are better understood.

In conclusion, we presented an automated tumor segmentation framework for colorectal cancer histology WSIs based on persistent homology. The proposed framework is validated on two independent datasets, consisting of both malignant cases and healthy cases. Extensive comparative analysis demonstrated better than state-of-the-art performance of the proposed algorithms. This study provides an insight into topological persistence of an image and may constitute the first step towards interpretable incorporation of homology features in the domain of histopathology image analysis. The proposed homology profiles model the intrinsic phenomena of cell connectivity and may be applicable to other similar problems in the computational pathology.

\section*{References}

\bibliography{PHP_MEDIA}

\end{document}